%% file: editor/templates/editor.tex
\documentclass[graybox, envcountchap]{svmult}

\usepackage{mathptmx}        % selects Times Roman as basic font
\usepackage{helvet}          % selects Helvetica as sans-serif font
\usepackage{courier}         % selects Courier as typewriter font
\usepackage{makeidx}         % allows index generation
\usepackage{graphicx}        % standard LaTeX graphics tool
\usepackage{multicol}        % used for the two-column index
\usepackage[bottom]{footmisc}% places footnotes at page bottom

\makeindex             % used for the subject index
\begin{document}

\frontmatter%%%%%%%%%%%%%%%%%%%%%%%%%%%%%%%%%%%%%%%%%%%%%%%%%%%%%%

\include{dedic}

\include{foreword}

\include{preface}

\include{acknow}

\tableofcontents
\include{cblist}
\include{acronym}

\mainmatter%%%%%%%%%%%%%%%%%%%%%%%%%%%%%%%%%%%%%%%%%%%%%%%%%%%%%%%
\include{part}

\include{author}

\backmatter%%%%%%%%%%%%%%%%%%%%%%%%%%%%%%%%%%%%%%%%%%%%%%%%%%%%%%%
\appendix
\include{appendix}
\include{glossary}
\printindex

%%%%%%%%%%%%%%%%%%%%%%%%%%%%%%%%%%%%%%%%%%%%%%%%%%%%%%%%%%%%%%%%%%%%%%

\end{document}

%% file: editor/templates/dedic.tex
%%%%%%%%%%%%%%%%%%%%%%% dedic.tex %%%%%%%%%%%%%%%%%%%%%%%%%%
%
% sample dedication
%
% Use this file as a template for your own input.
%
%%%%%%%%%%%%%%%%%%%%%%%% Springer %%%%%%%%%%%%%%%%%%%%%%%%%%

\begin{dedication}
Use the template \emph{dedic.tex} together with the Springer document class SVMono for monograph-type books or SVMult for contributed volumes to style a quotation or a dedication\index{dedication} at the very beginning of your book in the Springer layout
\end{dedication}

%% file: editor/templates/foreword.tex
%%%%%%%%%%%%%%%%%%%%%%foreword.tex%%%%%%%%%%%%%%%%%%%%%%%%%%%
% sample foreword
%
% Use this file as a template for your own input.
%
%%%%%%%%%%%%%%%%%%%%%%%% Springer %%%%%%%%%%%%%%%%%%%%%%%%%%

\foreword

Use the template \textit{foreword.tex} together with the Springer document class SVMono (monograph-type books) or SVMult (edited books) to style your foreword\index{foreword} in the Springer layout. 

The foreword covers introductory remarks preceding the text of a book that are written by a \textit{person other than the author or editor} of the book. If applicable, the foreword precedes the preface which is written by the author or editor of the book.

\vspace{\baselineskip}
\begin{flushright}\noindent
Place, month year\hfill {\it Firstname  Surname}\\
\end{flushright}

%% file: editor/templates/preface.tex
%%%%%%%%%%%%%%%%%%%%%%preface.tex%%%%%%%%%%%%%%%%%%%%%%%%%%%%%%%%%%%%%%%%%
% sample preface
%
% Use this file as a template for your own input.
%
%%%%%%%%%%%%%%%%%%%%%%%% Springer %%%%%%%%%%%%%%%%%%%%%%%%%%

\preface

Use the template \emph{preface.tex} together with the Springer document class SVMono (monograph-type books) or SVMult (edited books) to style your preface in the Springer layout.

A preface\index{preface} is a book's preliminary statement, usually written by the \textit{author or editor} of a work, which states its origin, scope, purpose, plan, and intended audience, and which sometimes includes afterthoughts and acknowledgments of assistance. 

When written by a person other than the author, it is called a foreword. The preface or foreword is distinct from the introduction, which deals with the subject of the work.

Customarily \textit{acknowledgments} are included as last part of the preface.

\vspace{\baselineskip}
\begin{flushright}\noindent
Place(s),\hfill {\it Firstname  Surname}\\
month year\hfill {\it Firstname  Surname}\\
\end{flushright}

%% file: editor/templates/acknow.tex
%%%%%%%%%%%%%%%%%%%%%%acknow.tex%%%%%%%%%%%%%%%%%%%%%%%%%%%%%%%%%%%%%%%%%
% sample acknowledgement chapter
%
% Use this file as a template for your own input.
%
%%%%%%%%%%%%%%%%%%%%%%%% Springer %%%%%%%%%%%%%%%%%%%%%%%%%%

\extrachap{Acknowledgements}

Use the template \emph{acknow.tex} together with the Springer document class SVMono (monograph-type books) or SVMult (edited books) if you prefer to set your acknowledgement section as a separate chapter instead of including it as last part of your preface.

%% file: editor/templates/cblist.tex
%%%%%%%%%%%%%%%%%%%%clist.tex %%%%%%%%%%%%%%%%%%%%%%%%
%                                                    
% sample list of contributors and their addresses    
%                                                    
% Use this file as a template for your own input.    
%                                                    
%%%%%%%%%%%%%%%%%%%%%%%% Springer %%%%%%%%%%%%%%%%%%%%
\contributors

\begin{thecontriblist}
Firstname Surname
\at ABC Institute, 123 Prime Street, Daisy Town, NA 01234, USA, \email{smith@smith.edu}
\and
Firstname Surname
\at XYZ Institute, Technical University, Albert-Schweitzer-Str. 34, 1000 Berlin, Germany, \email{meier@tu.edu}
\end{thecontriblist}

%% file: editor/templates/acronym.tex
%%%%%%%%%%%%%%%%%%%%%%acronym.tex%%%%%%%%%%%%%%%%%%%%%%%%%%%%%%%%%%%%%%%%%
% sample list of acronyms
%
% Use this file as a template for your own input.
%
%%%%%%%%%%%%%%%%%%%%%%%% Springer %%%%%%%%%%%%%%%%%%%%%%%%%%

\extrachap{Acronyms}

Use the template \emph{acronym.tex} together with the Springer document class SVMono (monograph-type books) or SVMult (edited books) to style your list(s) of abbreviations or symbols in the Springer layout.

Lists of abbreviations\index{acronyms, list of}, symbols\index{symbols, list of} and the like are easily formatted with the help of the Springer-enhanced \verb|description| environment.

\begin{description}[CABR]
\item[ABC]{Spelled-out abbreviation and definition}
\item[BABI]{Spelled-out abbreviation and definition}
\item[CABR]{Spelled-out abbreviation and definition}
\end{description}

%% file: editor/templates/part.tex
%%%%%%%%%%%%%%%%%%%%%part.tex%%%%%%%%%%%%%%%%%%%%%%%%%%%%%%%%%%
% 
% sample part title
%
% Use this file as a template for your own input.
%
%%%%%%%%%%%%%%%%%%%%%%%% Springer %%%%%%%%%%%%%%%%%%%%%%%%%%

\begin{partbacktext}
\part{Part Title}
\noindent Use the template \emph{part.tex} together with the Springer document class SVMono (monograph-type books) or SVMult (edited books) to style your part title page and, if desired, a short introductory text (maximum one page) on its verso page in the Springer layout.

\end{partbacktext}

%% file: author.tex
\newcommand{\jessica}[1]{\textcolor{magenta}{\bf\small [#1 --jessica]}}
\newcommand{\cathy}[1]{\textcolor{violet}{\bf\small [#1 --cathy]}}
\newcommand{\prakhar}[1]{\textcolor{blue}{\bf\small [#1 --prakhar]}}
\newcommand{\maxine}[1]{\textcolor{green}{\bf\small [#1 --maxine]}}

\title*{Understanding the Effectiveness of Very Large Language Models on Dialog Evaluation}
\titlerunning{Very Large Language Models for Dialog Evaluation}
% Use \titlerunning{Short Title} for an abbreviated version of
% your contribution title if the original one is too long
\author{Jessica Huynh, Cathy Jiao, Prakhar Gupta, Shikib Mehri, Payal Bajaj, Vishrav Chaudhary, Maxine Eskenazi}
\authorrunning{J. Huynh et al.}
% Use \authorrunning{Short Title} for an abbreviated version of
% your contribution title if the original one is too long
\institute{Jessica Huynh \at Carnegie Mellon University, \email{jhuynh@cs.cmu.edu}
\and Cathy Jiao \at Carnegie Mellon University, \email{cljiao@cs.cmu.edu}
\and Prakhar Gupta \at Carnegie Mellon University, \email{prakharg@cs.cmu.edu}
\and Shikib Mehri \at Amazon, \email{asmehri@amazon.com} (work done while at Carnegie Mellon University)
\and Payal Bajaj \at Microsoft Turing, \email{pabajaj@microsoft.com}
\and Vishrav Chaudhary \at Microsoft Turing, \email{vchaudhary@microsoft.com}
\and Maxine Eskenazi \at Carnegie Mellon University, \email{max@cs.cmu.edu}}
%
% Use the package "url.sty" to avoid
% problems with special characters
% used in your e-mail or web address
%

\maketitle

\input{templates/sections/0_Abstract}
\input{templates/sections/1_Introduction}
\input{templates/sections/2_RelatedWork}
\input{templates/sections/3_Datasets}
\input{templates/sections/4_BaselineTNLG}
\input{templates/sections/5_Conclusion}
\input{templates/sections/6_Acknowledgements}

% \scriptsize
\smaller
\bibliography{anthology, custom}
\bibliographystyle{styles/spbasic}

\input{templates/sections/7_Appendix}

%% file: templates/sections/0_Abstract.tex
\abstract{Language models have steadily increased in size over the past few years. They achieve a high level of performance on various natural language processing (NLP) tasks such as question answering and summarization. Large language models (LLMs) have been used for generation and can now output human-like text. Due to this, there are other downstream tasks in the realm of dialog that can now harness the LLMs' language understanding capabilities. Dialog evaluation is one task that this paper will explore. It concentrates on prompting with LLMs: BLOOM, OPT, GPT-3, Flan-T5, InstructDial and TNLGv2. The paper shows that the choice of datasets used for training a model contributes to how well it performs on a task as well as on how the prompt should be structured. Specifically, the more diverse and relevant the group of datasets that a model is trained on, the better dialog evaluation performs. This paper also investigates how the number of examples in the prompt and the type of example selection used   affect the model's performance.}

%% file: templates/sections/1_Introduction.tex
\section{Introduction}
\label{sec:intro}

In recent years, language models such as GPT-3~\cite{brown2020language} have grown larger, and their performance on downstream natural language processing (NLP) tasks has significantly improved in low-resource settings where only a few instances per task are available (few-shot).
The larger these models are, the higher their performances trend on tasks such as language generation and evaluation~\cite{wei2022emergent}. 
They can generate coherent, fluent and interesting responses. 
However, they can also produce responses that are repetitive and un-engaging \cite{roller-etal-2021-recipes}, in addition to being hard to control.
% However, they are hard to control and can also produce responses that are repetitive and un-engaging \cite{roller-etal-2021-recipes}.
Dialog evaluation is the task of assessing the quality of responses generated by dialog models in terms of properties like those mentioned above.
However, one significant impediment for open-domain dialog generation research is the lack of meaningful automatic metrics for open-domain dialog evaluation.
% The lack of meaningful automatic metrics for open-domain dialog evaluation is a significant impediment for open-domain dialog generation research. 
Standard language generation metrics have been shown to be ineffective for dialog evaluation~\cite{gupta-etal-2019-investigating}, a large part of which is because conversations can be followed by \textit{multiple valid} responses. Standard automatic metrics (e.g. BLEU \cite{papineni2002bleu}), which use references for evaluation, cannot deal with this quality, known as the \textit{one-to-many} response problem. 
Many recently introduced automatic metrics for dialog evaluation~\cite{mehri-eskenazi-2020-unsupervised, gupta-etal-2021-synthesizing} have attained increasingly stronger correlations with human judgment. 
Since human dialog evaluation typically measures multiple fine-grained properties (e.g. appropriate, interesting, consistent), automatic evaluation metrics should be expected to do the same.
This paper explores several fine-grained metrics that are measured both at turn-level (i.e. relevance and fluency), and dialog-level (i.e. consistency and coherence). 

Automatic dialog evaluation continues to be an evolving topic, but with fine-grained metrics and definitions varying across different human-annotated datasets \cite{mehri-eskenazi-2020-usr, zhao-etal-2020-designing}, it is important to be able to create reasonable automatic metrics with limited data.
Large language models (LLMs) that have been pre-trained on large-scale datasets are able to perform zero and few-shot inference~\cite{radford2019language, sanh2021multitask}, and they have exhibited good reasoning skills~\cite{brown2020language, wei2022emergent} in addition to having implicitly learned some notion of dialog quality~\cite{mehri-eskenazi-2020-unsupervised}. This makes them suitable for open-domain dialog evaluation in zero-shot and extreme few-shot settings. While there have been a few attempts to use LLMs for dialog evaluation~\cite{thoppilan2022lamda}, there has not, to our knowledge, been a systematic study of LLMs for this task. 
This paper explores several aspects of LLM use in dialog evaluation: the effect of model type and size and the choice of training data as well as the use of in-context examples for dialog evaluation (the number and quality of the examples used).
The experiments herein employ benchmarks to test both how well LLMs can be used for fine-grained evaluation, and how generalizable the models' performance is across multiple domains and datasets.
% Results show that although the much larger language models have superior performance, it is not as significant as could be expected from a language model of corresponding size. This paper delves into the differences in performance, the reasons that may be the cause and proposes what can be done to improve LLM performance.

% why we are not getting much better performance - what it's trained on
% what specifically can be done so LLMs can achieve the better performance we are expecting

%% file: templates/sections/2_RelatedWork.tex
\section{Related Work}
\label{sec:related_work}

\subsection{LLMs}
\label{sec:llm}

\begin{figure}[h]
\centering
\includegraphics[width=0.67\textwidth]{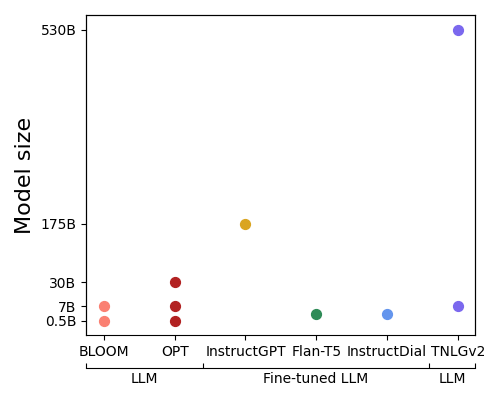}
\caption{Large Language Models, comparison of select approximate sizes}
\label{fig:modelsize}
\end{figure}

Several LLMs have been released recently: T5 \cite{2020t5}, GPT-3 \cite{brown2020language}, BLOOM \cite{bigscience_workshop_2022}, OPT \cite{zhang2022opt}, and TNLGv2 \cite{smith2022using}. The following models, the sizes of which are shown in Figure \ref{fig:modelsize}, are explored here:

\begin{itemize}
    \item T5, trained on the 750B Colossal Clean Crawled Corpus (C4) contains heuristically cleaned natural language English text from the web. Specific models considered are:
    \begin{itemize}
        \item Flan-T5 \cite{chung2022scaling}, T5 fine-tuned on 1836 tasks, including dialog tasks and data. 
        \item InstructDial \cite{gupta2022improving}, T5 fine-tuned specifically on 48 dialog tasks. 
    \end{itemize}
    \item GPT-3 includes a 570B filtered CommonCrawl corpus \cite{2020t5} in addition to WebText \cite{radford2019language}, Books1, Books2, and Wikipedia \cite{books2020}. 
    \begin{itemize}
        \item InstructGPT (text-davinci-002) \cite{ouyang2022training}, GPT-3 fine-tuned with a prompting \newline dataset and 175B parameters. 
    \end{itemize}
    \item BLOOM was trained on 46 languages and 13 programming languages with a multilingual focus. 
    \item OPT contains data from the RoBERTa corpus \cite{liu2019roberta}, the Pile \cite{pile}, and PushShift.io Reddit \cite{baumgartner2020pushshift, roller-etal-2021-recipes}. 
    \item TNLGv2 is trained on a subset of the Pile (notably excluding corpora classified as having natural dialog), two CommonCrawl snapshots \cite{2020t5}, RealNews \cite{realnews2019}, and CC-Stories \cite{cc2018}.
\end{itemize}

As the number of parameters in these models increases, performance also increases: TNLGv2 530B, with around three times the number of parameters, outperforms the original GPT-3 on a variety of NLP tasks \cite{smith2022using}. LLMs are also generalizable; they perform well on many NLP tasks in few-shot settings and zero-shot settings~\cite{wei2021finetuned, sanh2021multitask}. 
% OPT-175B, \cite{zhang2022opt}
% The amount of resources used to train LLMs
% In addition, although similarly-sized LLMs such as OPT are developed with less resources than GPT-3, they show comparable few-shot and zero-shot performance. 
However, several drawbacks and areas for exploration remain for LLMs that should be noted.
Recent work has shown that performance on certain zero-shot tasks plateaus as model parameter size grows exponentially \cite{brown2020language}.
LLMs also struggle with parsing social situations \cite{sap2022neural} and correctly using context \cite{agarwal2021interpretability}, which are important in dialog settings.
This raises questions on the performance of LLMs for dialog evaluation, and how an LLM's performance changes as it increases in size.

The data that a model is trained on also influences the performance of downstream tasks. T5 is fine-tuned on various subtasks, but pre-trained with C4. When pre-trained with domain-specific data, T5 performs better on tasks in that domain \cite{beltagy2019scibert, 2020t5}. Furthermore, adding several domains of data during pre-training makes the model likely to perform better \cite{liu2019roberta, zhang2022opt, chowdhery2022palm}. Notably,  BLOOM, OPT, Flan-T5, InstructGPT, and InstructDial are partially trained or fine-tuned on dialog datasets. Details on the content of these datasets can be found in Appendix \ref{sec:llmdata}. This is important because natural dialog data is difficult to obtain, so either scripted conversations or Reddit threads are used since they are the most readily available. This dearth of data is the reason that few-shot prompting is of interest. While work such as \cite{wei2022emergent} acknowledges emergent abilities in larger language models in few-shot prompting settings, this paper explores discrepancies in performance specifically for dialog evaluation. 

\subsection{Dialog Evaluation}
\label{sec:dialogeval}
Dialog evaluation presents a unique combination of challenges; it must consider multiple speakers \cite{zhang2021dialoguebert}, context that informs the current dialog turn, and the one-to-many aspect mentioned above \cite{zhao2017learning}. 

Metrics such as USR \cite{mehri-eskenazi-2020-usr} and FED \cite{mehri-eskenazi-2020-unsupervised} were created to address some of these challenges; they are reference-free, capture complex aspects of dialog, and have good correlation with human evaluation. These metrics use models such as RoBERTa (125 million parameters) \cite{liu2019roberta} and DialoGPT (345 or 762 million parameters) \cite{zhang2019dialogpt} respectively.
However, the best performing versions of these models are smaller than most models examined in this paper, and are fine-tuned on dialog data or on a specific dialog task.
Other automatic evaluation metrics include GRADE \cite{huang-etal-2020-grade} and DEB \cite{sai2020improving}. With current LLMs' large increase in hyperparameters, their plethora of training data, and their promising generalizable performance on NLP tasks, these model-based metrics should improve as well.

% Metrics such as USR \cite{mehri-eskenazi-2020-usr} and FED \cite{mehri-eskenazi-2020-unsupervised} were created to be  reference-free, capture complex aspects of dialog, and have good correlation with human evaluation. These metrics use models such as RoBERTa (125 million parameters) \cite{liu2019roberta} and DialoGPT (345 or 762 million parameters) \cite{zhang2019dialogpt} respectively, smaller than most models examined in this paper. The best performing versions of these models were fine-tuned on dialog data, or on a specific dialog task. Other automatic evaluation metrics include: GRADE \cite{huang-etal-2020-grade} and DynaEval \cite{zhang-etal-2021-DynaEval}.
% % Other automatic evaluation metrics include: MAUDE \cite{sinha-etal-2020-learning}, GRADE \cite{huang-etal-2020-grade}, FlowScore \cite{li-etal-2021-conversations}, USL-H \cite{phy-etal-2020-deconstruct}, QuestEval \cite{scialom-etal-2021-questeval}, DynaEval \cite{zhang-etal-2021-DynaEval}, and DialogRPT \cite{gao-etal-2020-dialogue}. 
% With the advent of LLMs and their promising generalizable performance on NLP tasks, these metrics should improve as well.

% TNLGv2 is not trained or fine-tuned on such data, but with the exponential increase in parameters, we hypothesize that there will be large gains in these metrics in the few-shot setting. 

\subsection{Example selection for few-shot learning}

The example selection process for prompting LLMs is of great interest. Prompting an LLM with a task and a few examples enables the model to adapt to a new task without completely fine-tuning it. 
% In-context examples allow LLMs to figure out the context of the task, and then make predictions using this information.
In particular, in-context examples can provide important cues to help LLMs make predictions on tasks.
% Prompts can be automatically generated \cite{reynolds2021prompt, gao2020making, zhou2022learning}, or selected from existing data. 
% \jessica{wording, more about what it is?} 
% algorithmically chosen examples
Recent work has used a variety of methods to examine example selection. Common methods measure semantic similarity between example embeddings \cite{liu2021makes, su2022}. Alternatively, retrieval methods (e.g. BM25 \cite{bm25})
% and BRUTEFORCE \cite{bruteforce} 
have been used directly, or as a precursor to training a selection retriever \cite{rubin2021}. 

These example selection methods have shown promise in few-shot NLP tasks. In \cite{su2022}, the two-step framework for annotating and selecting in-context examples from large unlabeled data showed competitive performance across 10 tasks such as classification, commonsense reasoning, dialog state tracking, and code generation. \cite{liu2021makes} showed that selecting examples with similar sentence embeddings yields higher GPT-3 performance than random selection. However, the authors acknowledge that further investigation is required to find more efficient in-context example retrieval methods. 

Moreover, the wording and order of examples presented in prompts can also affect model performance~\cite{gao2020making, liu2021makes, jiang2020can}. \citet{fantastic2021} observed order sensitivity across 0.1B to 175B parameter GPT-2 and GPT-3 models when the models were probed with different text classification tasks and several in-context examples. Also, the wording of the in-context examples depends on the data used for model training; for unfamiliar prompt formats, model performance may decrease \cite{jiang2020can}. Increasing the size of the model and the amount of data does not resolve the issue since the same instability is still prevalent \cite{zhao2021calibrate}. Thus this paper studies the effect of example selection on dialog evaluation. 

%% file: templates/sections/3_Datasets.tex
\section{Evaluation Settings}
\label{sec:datasets}
Two settings for dialog evaluation are explored: fine-grained evaluation and multi-domain evaluation.  In-context examples are explored in both.

\subsection{Fine-Grained Evaluation}

Fine-grained metrics can be measured at both the turn level (e.g. informativeness and relevance), and the dialog level (e.g. coherence and diversity). The FED dataset \cite{mehri-eskenazi-2020-unsupervised} is used. It consists of 124 open-domain dialogs of humans with humans or with machines, for which each dialog has 3 responses that are chosen for annotation (8 turn-level and 10 dialog-level qualities along with overall turn- and dialog-level quality). This dataset was chosen due to the large number of previously studied fine-grained qualities as listed in Section \ref{fed}, with the exception of correctness and error recovery, which are only specifically present in FED. 

In the experiments, the LM is prompted to output a rating (an integer value - see Appendix \ref{sec:exfed}) to evaluate each fine-grained quality in a response.
The final rating for each fine-grained quality is a weighted sum of the $K$-top ratings outputted from the LM. Formally, given the $K$-top predicted ratings ${r_1, r_2, ..., r_K}$ along with their corresponding log probabilities, ${p_1, ...,p_K}$, the weight, $w_i$, of each rating $r_i$ is derived as:

$$w_i = \frac{p_i}{\sum_{j=1}^{K} p_j}$$

The final rating, $r$, is calculated as:

$$r = \sum_{i=1}^{K} r_i * w_i$$

In order to provide a more accurate view of the LM's performance, $K = 3$ in the following experiments. Additionally, this scoring mechanism converts the LM predictions onto a continuous scale, which more closely mirrors the average of human ratings. Results are reported with the Spearman correlations to the average human ratings for each fine-grained quality.

% We use the $K$-top ratings instead of the best rating to take into account wavering judgements for dialog evaluation and synthesize the LM's best predictions.

% For example, annotators may have wavered between two scores.

% annotated that were mostly consolidated from previous work.

% These include a subset of the USR dataset \cite{mehri2020usr}, which consists of knowledge-grounded open-domain conversations with 60 dialog contexts with 6 responses to each context annotated on 6 qualities

\subsection{Multi-domain Evaluation}
This task tests automatic dialog evaluation metrics for robustness across multiple dialog domains. The analysis uses only the overall quality metric since many of the domain datasets do not have fine-grained annotations. The Spearman correlation is used between human ratings and model predictions on the evaluation sets released by DSTC 10 Track 5~\cite{chen2021automatic} ``Automatic Evaluation and Moderation of Open-domain Dialogue Systems''. These sets contain human judgement ratings for dialog responses. In this setting, a model is shown a dialog context and a response, and it outputs ``yes'' if the response is a good response to that context, otherwise it outputs ``no''. An example can be seen in Appendix \ref{sec:exdstc10}.
% We calculate the probability of ``yes'' or that the response is relevant to the context as $p(yes) = p(yes)/(p(yes)+p(no))$.
The probability of the ``goodness" of the response (i.e., the rating), $g$, is calculated as:

$$g = \frac{p_{model}(yes)}{p_{model}(yes)+p_{model}(no)}$$

where $p_{model}(yes)$ and $p_{model}(no)$ are the log probilities of the model outputs for ``yes" and ``no". Evaluation is carried out on 8 representative evaluation sets out of the 14 DSTC10 evaluation sets~\cite{chen2021automatic}. This subset was chosen because it covers multiple domains and datasets, such as persona, topic and chitchat-based responses. A robust dialog metric should perform well across all the domains and evaluation sets considered.

The evaluation sets used for fine-grained evaluation, FED-Turn (FT) and FED-Dial (FD) \cite{mehri-eskenazi-2020-unsupervised}, are included as two of the eight datasets. The other datasets include: TopicalChat-USR (TU, knowledge-grounded open-domain conversations rated for six different dialog qualities) \cite{mehri-eskenazi-2020-usr}; PersonaChat-USR (PU, persona-conditioned conversations annotated with the USR schema) \cite{mehri-eskenazi-2020-usr}; DailyDialog-Zhao (DZ, more formal language conversations rated for appropriateness) \cite{zhao-etal-2020-designing}; DailyDialog-Gupta (DGU, rated for appropriateness) \cite{gupta-etal-2019-investigating}; DailyDialog-GRADE (DGR; annotated for coherence) \cite{huang-etal-2020-grade}; and Empathetic-GRADE (EG, emotionally grounded conversations annotated for coherence) \cite{huang-etal-2020-grade}. 
Although some of these datasets are not directly annotated for whether a response is good, the metric they use remains a component for overall quality, and thus it is treated as the indicator of the overall quality of the response in the experiments.

\subsection{In-Context Examples}

This paper uses two methods for example selection: random selection, and algorithmic selection using BM25 \cite{manning2008introduction} which calculates document similarity. The examples remain consistent for each evaluation test point. The random selection experiment is run three times, and the mean and standard deviation of the runs are reported. There are three configurations for BM25 between the test point and each possible example point - comparing the context only ($\textsc{BM25}_\textsc{c}$), the response only ($\textsc{BM25}_\textsc{r}$), and the concatenated context and response together ($\textsc{BM25}_\textsc{c+r}$).

With the FED dataset, an additional method, manual selection, is added for example selection. For each fine-grained dialog quality, a set of three dialogs which span a wide range of ratings is chosen that remains constant over every test point. In theory, the model should be able to show increased performance if it sees examples of very good, good and bad responses for fine-grained metrics.
For the DSTC10 datasets, an additional experiment tested how the number of examples used affects model performance. 
% Since overall quality metrics consolidate many factors together, we hypothesize that a model may need more examples to capture the quality compared to fine-grained metrics that are more specific.

% our first set of experiments are on topicalchat-usr and fed-turn, fed-dialog

% second set is those + personachat, dailydialog, empathetic

% topical chat, personachat (2), dailydialog (2), empathetic (1), fed (2)

% TopicalChat-USR (TU) - same as the one i used, 6 responses per context

% context + fact, knowledge-grounded, open-domain; collected in AMT

% rated for understandable, natural, maintains context, interesting, uses knowledge, overall quality

% PersonaChat-USR (PU)

% AMT; persona-conditioned

% rated for understandable, natural, maintains context, interesting, uses knowledge, overall quality -> we use overall quality

% PersonaChat-Zhao (PZ)

% rated for appropriateness

% DailyDialog-Gupta (DGU)

% dailydialog - more formal, on 'English learner to practice English dialog in daily life', 'topic focus'

% annotated for appropriateness

% DailyDialog-GRADE (DGR)

% context response pairs, 2 false responses

% graded on coherence
 
% Empathetic-GRADE (EG)

% emotion label grounded on AMT

% graded on coherence

% FED-Turn (FT)

% b/t humans + systems, humans + humans, open domain

% turn level quality

% FED-Dial (FD)

% dialog level quality

% USR - 60 dialogues, 360 responses
% % https://aclanthology.org/2020.acl-main.64.pdf

% FED - 124 dialogues, 3 system responses throughout the conversation
% % https://arxiv.org/pdf/2006.12719.pdf

%% file: templates/sections/4_BaselineTNLG.tex
\section{Experiments and Results}
\label{sec:baseline}

% \jessica{put this here or in the previous section with in-context examples?} 
The in-context example experiments are carried out on the largest available model, 530B TNLGv2, to explore the ceiling of model performance on the dialog evaluation task. 6.7B TNLGv2 is used for a direct comparison of how much performance gain is provided by using more parameters. 

BLOOM and OPT are examined up to 7B and 30B respectively for the fine-grained metric evaluation task. \footnote{Due to limitations in compute power, larger BLOOM and OPT models were not explored. However, as the largest available GPT-3 model is explored, the comparisons appear sufficient to show the performance of a variety of LLMs.} Smaller LLMs do not perform as well with in-context examples unless they have been specifically tuned for the task, so only the 7B and 6.7B models for BLOOM and OPT respectively are explored for the DSTC10 datasets. Flan-T5 and InstructDial are analyzed in the 3B setting for consistency. Lastly, InstructGPT (text-davinci-002) is used, which has 175B parameters.

% BLOOM and OPT are examined up to 7B and 30B respectively and, FlanT5 up to 3B due to constraints on processing power. The largest version of InstructDial was used, with 3B parameters, and GPT-3, with 175B parameters.

\subsection{Fine-grained Metric Evaluation} \label{fed}
FED is separated into turn-level and dialog-level metrics. The dataset has annotations for 8 different turn-level metrics, consisting of \textit{interestingness}, \textit{engagingness}, \textit{specificity}, \textit{relevance}, \textit{correctness}, \textit{semantic appropriateness}, \textit{understandability}, and  \textit{fluency}, with the addition of \textit{overall quality}. FED annotates three different responses for each dialog context; one FED dialog is treated as one example. The corresponding rating is inserted after the response statement in the prompt, an example of which can be seen in Appendix \ref{sec:exfed}. FED also looks at 10 different dialog-level metrics for a system's responses: \textit{coherence}, \textit{error recovery}, \textit{consistency}, \textit{diversity}, \textit{topic depth}, \textit{likeability}, \textit{understandingness}, \textit{flexibility}, \textit{informativeness}, and  \textit{inquisitiveness}, with \textit{overall quality} included. The model is prompted with the full dialog context with the rating.

The FED metric was previously evaluated with both fine-tuned (ft) and from-scratch 345M and 762M DialoGPT \cite{zhang2019dialogpt} models. In the following experiments on FED, 3 in-context examples were used for prompting in Tables \ref{table:fedturn}, \ref{table:feddialog}, \ref{table:llmturn} and \ref{table:llmdialog} and Appendix \ref{sec:fedfull} and \ref{sec:fedmodels}. 

\subsubsection{In-Context Example Selection}
\label{sec:incontext}
This setting evaluates 2 versions of the TNLGv2 model: 6.7B and 530B. 
These models are compared to the 762M ft DialoGPT model and the results are shown in Tables \ref{table:fedturn} and \ref{table:feddialog} and Appendix \ref{sec:fedfull}. 

First, the performances of these models are compared over the three example selection methods: manual, random, and algorithmic.
With manually chosen in-context examples, the 530B TNLGv2 model outperforms the DialoGPT model on almost all turn-level metrics except for \textit{understandability} and \textit{fluency}. There are significant gains in all of the dialog-level metrics as well.
Since DialoGPT is fine-tuned on Reddit threads, more casual language is expected, compared to models like TNLGv2 where many of the training datasets consist of more formal language. Since the wording of conversational responses tends to be more casual, it is not surprising that the fine-tuned DialoGPT model outperforms even the largest TNLGv2 model for \textit{fluency} and \textit{understandability}. However, the TNLGv2 models show large improvement on predicting \textit{turn-} and \textit{dialog-level quality}.
% Notably, the overall turn-level and dialog-level metrics show large improvement. 
This suggests that the TNLGv2 models have a strong grasp on overall quality, which may be due to training on more formal language. 

$\textsc{BM25}_\textsc{c+r}$ generally outperforms $\textsc{BM25}_\textsc{c}$ and $\textsc{BM25}_\textsc{r}$.
However, when choosing examples with $\textsc{BM25}_\textsc{c+r}$, the correlation of \textit{understandability} with human annotations increases significantly when using the 6.7B TNLGv2 model. 6.7B TNLGv2 consistently outperforms 530B TNLGv2 in this aspect with any BM25 method. It appears that the smaller model is more influenced by the similarity of language in the examples than the larger one.

Even when given random examples, the TNLGv2 models outperform the 762M ft DialoGPT model on a majority of the fine-grained metrics. This shows that larger models can better detect what constitutes a good response based on these metrics even if they are not given hand-picked examples. However, they generally do not outperform the manually or algorithmically chosen examples as expected.

An additional observation is that there are certain factors that cause models to perform better or worse on specific metrics: number of parameters the model has, the type of training data, and the difficulty of the task. LLMs are able to provide increases in performance of over 50\% for 15 out of 20 turn- and dialog-level metrics compared to DialoGPT with 530B TNLGv2 and manually-chosen examples. However, if the 530B TNLGv2 model is compared to the 6.7B TNLGv2 model, this increase is only observed for 2 out of the 20 metrics: \textit{correctness} and \textit{understandability}. LLMs can achieve high correlations with human judgement, but there is a limit to how much more performance gains can increase with extremely large models.

\textit{Specificity}, \textit{relevance}, and \textit{correctness} all relate to the context of the conversation while the other metrics are more turn-specific. It follows that \textit{relevance} and \textit{correctness} with $\textsc{BM25}_\textsc{c+r}$ on the 6.7B TNLGv2 model outperform the 530B TNLGv2 model with manual examples. However, \textit{specificity} performs worse. Choosing both diverse ratings and similar example points are important. This finding further supports the idea that the nature of the data used to train these LLMs is important. Had the training data been more similar to conversational language, an increase could have been observed in the correlations for these metrics without choosing algorithmically similar examples.

TNLGv2 struggles with \textit{understandability}; it performs the worst at the highest correlation of 0.193. It also has unstable performance; performing at significance with random examples and with algorithmically chosen examples on 6.7B, but not with manually chosen ones. This shows that choosing examples with diverse ratings helps a model less for metrics that it already performs poorly on; it would better benefit from examples that are similar.

In general, even with the difference in training data, it is easier to obtain an overall sense of the conversation than a metric for a single turn for the larger models due to the large amount of parameters and variety of data that they have seen. When choosing examples based on context, the larger models generally perform worse; it appears that having different examples is more important for dialog-level metrics than for turn-level metrics.

\begin{table*}[t]
\begin{center}
\scriptsize
\centering
\setlength{\tabcolsep}{4pt}
% \begin{tabular}{ lc|cc|cc|cc } 
\begin{tabular}{ lccccccc } 
  \toprule
 & & \multicolumn{2}{c}{manual} & \multicolumn{2}{c}{random} & \multicolumn{2}{c}{$\textsc{BM25}_\textsc{c+r}$} \\
 \cmidrule(lr){3-4}\cmidrule(lr){5-6}\cmidrule(lr){7-8}
%  \hline
 Quality & 762M ft & 6.7B & 530B & 6.7B & 530B & 6.7B & 530B \\
 \hline
 Interesting & 0.408 & 0.455 & \textbf{0.474} & 0.293 {\tiny $\pm$ 0.03} & 0.398{\tiny $\pm$ 0.02} & 0.358 & 0.383 \\
 Engaging & 0.318 & 0.459 & \textbf{0.484} & 0.235{\tiny $\pm$ 0.04} & 0.352{\tiny $\pm$ 0.02} & 0.378 & 0.383 \\
 Specific & 0.267 & 0.305 & \textbf{0.450} & 0.188{\tiny $\pm$ 0.02} & 0.289{\tiny $\pm$ 0.01} & 0.268 & 0.322 \\
 Relevant & 0.152 & 0.214 & 0.300 & 0.179{\tiny $\pm$ 0.04} & 0.299{\tiny $\pm$ 0.03} & \textbf{0.392} & 0.357\\
 Correct & 0.133 & 0.195 & 0.393 & 0.171{\tiny $\pm$ 0.04} & 0.338{\tiny $\pm$ 0.04} & \textbf{0.399} & 0.377 \\
 Sem. Approp. & 0.155 & 0.292 & \textbf{0.395} & 0.163{\tiny $\pm$ 0.03} & 0.270{\tiny $\pm$ 0.01} & 0.291 & 0.294 \\
 Understandable & 0.111 & 0.021* & 0.036* & 0.146{\tiny $\pm$ 0.02} & 0.129{\tiny $\pm$ 0.02} & \textbf{0.193} & 0.062* \\
 Fluent & \textbf{0.224} & 0.164 & 0.195 & 0.052*{\tiny $\pm$ 0.03} & 0.112*{\tiny $\pm$ 0.01} & 0.096* & 0.178 \\
 Overall & 0.209 & 0.371 & 0.475 & 0.256{\tiny $\pm$ 0.02} & 0.380{\tiny $\pm$ 0.01} & 0.474 & \textbf{0.514} \\
 \bottomrule
\end{tabular}
\caption{Turn-level fine-grained metrics on the FED dataset for manually, randomly, and \textsc{BM25} chosen examples over the TNLGv2 6.7B and 530B models. $\textsc{BM25}_\textsc{c+r}$ stands for examples chosen by BM25 considering both the context and the response of the test point. \textbf{Bold} values indicate the best value for the metric and * values indicate correlations that are not statistically significant.}
\label{table:fedturn}
\end{center}
\end{table*}

\begin{table*}[t]
\begin{center}
\scriptsize
\centering
\setlength{\tabcolsep}{5pt}
\begin{tabular}{ lccccccc } 
  \toprule
 & & \multicolumn{2}{c}{manual} & \multicolumn{2}{c}{random} & \multicolumn{2}{c}{$\textsc{BM25}_\textsc{c}$} \\
 \cmidrule(lr){3-4}\cmidrule(lr){5-6}\cmidrule(lr){7-8}
 Quality & 762M ft & 6.7B & 530B & 6.7B & 530B & 6.7B & 530B \\
 \hline
 Coherent & 0.251 & 0.599 & \textbf{0.727} & 0.443{\tiny $\pm$ 0.03} & 0.533{\tiny $\pm$ 0.02} & 0.618 & 0.512 \\
 Error Recovery & 0.165* & 0.474 & \textbf{0.578} & 0.348{\tiny $\pm$ 0.04} & 0.463{\tiny $\pm$ 0.06} & 0.492 &  0.419 \\
 Consistent & 0.116* & 0.276 & \textbf{0.382} & 0.270{\tiny $\pm$ 0.02} & 0.205* {\tiny $\pm$ 0.04} & 0.238 & 0.046* \\
 Diverse & 0.420 & \textbf{0.625} & 0.620 & 0.434{\tiny $\pm$ 0.06} & 0.490{\tiny $\pm$ 0.02} & 0.496 & 0.548 \\
 Topic Depth & 0.476 & 0.640 & \textbf{0.659} & 0.361{\tiny $\pm$ 0.03} & 0.531{\tiny $\pm$ 0.04} & 0.559 & 0.472 \\
 Likeable & 0.262 & 0.619 & \textbf{0.686} & 0.511{\tiny $\pm$ 0.03} & 0.580{\tiny $\pm$ 0.01} & 0.568 & 0.515 \\
 Understanding & 0.306 & 0.517 & \textbf{0.638} & 0.479{\tiny $\pm$ 0.06} & 0.496{\tiny $\pm$ 0.02} & 0.567 &  0.428 \\
 Flexible & 0.293 & 0.617 & \textbf{0.656} & 0.491{\tiny $\pm$ 0.05} & 0.553{\tiny $\pm$ 0.03} & 0.614 & 0.451 \\
 Informative & 0.288 & \textbf{0.569} & 0.547 & 0.391{\tiny $\pm$ 0.04} & 0.452{\tiny $\pm$ 0.04} & 0.523 & 0.419 \\
 Inquisitive & 0.163 & \textbf{0.537} & 0.527 & 0.436{\tiny $\pm$ 0.05} & 0.444{\tiny $\pm$ 0.02} & 0.334 & 0.252 \\
 Overall & 0.443 & 0.630 & \textbf{0.688} & 0.479{\tiny $\pm$ 0.05} & 0.570{\tiny $\pm$ 0.02} & 0.607 & 0.531 \\
 \hline
\end{tabular}
\caption{Dialog-level fine-grained metrics on the FED dataset for manually, randomly, and \textsc{BM25} chosen examples over the TNLGv2 6.7B and 530B models. $\textsc{BM25}_\textsc{c}$ stands for examples chosen by BM25 considering only the context of the test point.}
\label{table:feddialog}
\end{center}
\end{table*}

\subsubsection{Comparisons Across LLMs}

These model comparisons are performed using manually chosen in-context examples, since that is what generally performed the best in both turn-level and dialog-level metrics in Tables \ref{table:llmturn} and \ref{table:llmdialog}. Comparisons across smaller versions of BLOOM and OPT can be found in Appendix \ref{sec:fedmodels}.

Even though the large versions of BLOOM and OPT could not be run, it is apparent that both of these models outperform TNLGv2 on \textit{understandability}, and that OPT 6.7B can outperform TNLGv2 530B on \textit{fluency}. Data dissimilarities were noted above in Section \ref{sec:incontext} between the TNLGv2 model and the FED data. Although BLOOM was only trained on some English data, it has still seen some casual language, while OPT was partially trained on Reddit data. Thus the language appearing in the BLOOM and OPT training sets more closely matches that of the conversations used here. This explains the increase in performance. 

BLOOM 7B outperforms 6.7B TNLGv2 on \textit{correctness}, while OPT 6.7B outperforms 6.7B TNLGv2 on \textit{relevance}, \textit{correctness}, \textit{semantic appropriateness} and \textit{fluency} in addition. As previously noted, \textit{relevance} and \textit{correctness} are turn-level metrics that take more of the context into account, so with training data that is more similar to casual language, these models perform better. It should be noted that the \textit{overall turn-} and \textit{dialog-level quality} results were not surpassed by any smaller model, thus the very large models will have an advantage for overall metrics.

\begin{table*}[t]
\begin{center}
\scriptsize
\centering
\setlength\tabcolsep{4pt}
\begin{tabular}{ lcccccccc } 
  \toprule
 & \multicolumn{2}{c}{TNLG} & \multicolumn{1}{c}{BLOOM} & \multicolumn{2}{c}{OPT} & \multicolumn{1}{c}{Flan-T5} & \multicolumn{1}{c}{InstructGPT} \\
 \cmidrule(lr){2-3}\cmidrule(lr){4-4}\cmidrule(lr){5-6}\cmidrule(lr){7-7}\cmidrule(lr){8-8}
 Quality & 6.7B & 530B & 7B & 6.7B & 30B & 3B & 175B \\
 \hline
 Interesting & 0.455 & 0.474 & 0.291 & 0.429 & 0.399 & 0.519 & \textbf{0.551} \\
 Engaging & 0.459 & 0.484 & 0.435 & 0.446 & 0.349 & 0.425 & \textbf{0.489} \\
 Specific & 0.305 & \textbf{0.450} & 0.296 & 0.275 & 0.207 & 0.433 & 0.421 \\
 Relevant & 0.214 & 0.300 & 0.109 & 0.272 & 0.289 & 0.435 & \textbf{0.471} \\
 Correct & 0.195 & \textbf{0.393} & 0.235 & 0.342 & 0.354 & 0.378 & 0.376 \\
 Sem. Approp. & 0.292 & \textbf{0.395} & 0.258 & 0.371 & 0.382 & 0.277 & 0.374 \\
 Understandable & 0.021* & 0.036* & 0.159 & 0.131 & 0.073* & 0.297 & \textbf{0.382} \\
 Fluent & 0.164 & 0.195 & 0.111 & 0.201 & 0.188 & 0.200 & \textbf{0.204} \\
 Overall & 0.371 & 0.475 & 0.274 & 0.368 & 0.433 & 0.445 & \textbf{0.536} \\
 \hline
\end{tabular}
\caption{Turn-level fine-grained metrics on the FED dataset for manually chosen examples over the TNLGv2, BLOOM, OPT, Flan-T5, and InstructGPT models.}
\label{table:llmturn}
\end{center}
\end{table*}

\begin{table*}[t]
\begin{center}
\scriptsize
\centering
\setlength\tabcolsep{4pt}
\begin{tabular}{ lccccccc } 
  \toprule
 & \multicolumn{2}{c}{TNLG} & \multicolumn{1}{c}{BLOOM} & \multicolumn{2}{c}{OPT} & \multicolumn{1}{c}{FLAN-T5} & \multicolumn{1}{c}{InstructGPT}\\
 \cmidrule(lr){2-3}\cmidrule(lr){4-4}\cmidrule(lr){5-6}\cmidrule(lr){7-7}\cmidrule{8-8}
 Quality & 6.7B & 530B & 7B & 6.7B & 30B & 3B & 175B \\
 \hline
 Coherent & 0.599 & 0.727 & 0.613 & 0.558 & 0.584 & \textbf{0.730} & 0.707 \\
 Error Recovery & 0.474 & \textbf{0.578} & 0.474 & 0.377 & 0.479 & 0.398 & 0.560 \\
 Consistent & 0.276 & 0.382 & 0.323 & 0.237 & 0.309 & 0.410 & \textbf{0.517} \\
 Diverse & 0.625 & 0.620 & 0.498 & 0.454 & 0.607 & 0.544 & \textbf{0.628} \\
 Topic Depth & 0.640 & 0.659 & 0.637 & 0.544 & 0.609 & 0.650 & \textbf{0.680} \\
 Likeable & 0.619 & \textbf{0.686} & 0.566 & 0.544 & 0.571 & 0.659 & 0.672 \\
 Understanding & 0.517 & 0.638 & 0.484 & 0.505 & 0.483 & 0.637 & \textbf{0.694} \\
 Flexible & 0.617 & 0.656 & 0.499 & 0.528 & 0.592 & 0.595 & \textbf{0.688} \\
 Informative & 0.569 & 0.547 & 0.462 & 0.497 & 0.522 & \textbf{0.662} & 0.647 \\
 Inquisitive & 0.537 & 0.527 & 0.539 & 0.461 & 0.537 & 0.487 & \textbf{0.578} \\
 Overall & 0.630 & 0.688 & 0.531 & 0.374 & 0.530 & 0.585 & \textbf{0.690} \\
 \hline
\end{tabular}
\caption{Dialog-level fine-grained metrics on the FED dataset for manually chosen examples over the TNLGv2, BLOOM, OPT, Flan-T5, and InstructGPT models.}
\label{table:llmdialog}
\end{center}
\end{table*}

Flan-T5 outperforms the largest model, TNLGv2 530B, on \textit{interestingness}, \textit{relevance}, and \textit{understandability} at turn level and \textit{coherence}, \textit{consistency}, and \textit{informativeness} at dialog level. There is a larger performance drop for the \textit{semantic appropriateness}, \textit{error recovery}, and \textit{overall dialog-level quality} metrics. \textit{Error recovery} is a relatively new metric \cite{mehri-eskenazi-2020-unsupervised}. Even though Flan-T5 was fine-tuned on many dialog tasks, it may not have seen data that addresses this specific metric. Flan-T5 only has 3B parameters, and the fact that it outperforms 530B TNLGv2 shows the importance of use of dialog data during pre-training or fine-tuning.

% Before 530B TNLGv2 existed, GPT-3 was one of the largest language models available, at 175B parameters. 
InstructGPT, being fine-tuned with prompting at 175B parameters, is more suitable for the present experiments. It performs very well on both turn- and dialog-level metrics, outperforming 530B TNLGv2 on almost all metrics. Since InstructGPT has already seen prompting, the model can better understand a task through only instructions or combinations of instructions and in-context examples.

\subsection{DSTC10 Datasets} \label{dstc10}

The same set of experiments were carried out on the 8 datasets in the DSTC10 challenge in Tables \ref{tab:eval530b} and \ref{tab:evalother}, and Appendix \ref{sec:dstc106.7B}. The previous best performing metrics on DSTC10 are compiled from \cite{gupta2022improving}, which include both reference-free and fine-tuned metrics (see Appendix \ref{sec:dstc10}). Quality is evaluated in terms of how good a response is to the context. 

\subsubsection{In-Context Example Selection}

\begin{table*}[t]
% \small
% \footnotesize
\scriptsize
\centering
\setlength\tabcolsep{1.0pt}
\begin{tabular}{lcccccccccccc}
    \hline
   Model & TU & DZ & PU & DGU & DGR & FT & EG & FD \\
   \hline
   \rowcolor{Gray}
\multicolumn{9}{l}{\textit{Experiments with Random Examples}}
 \\
4ex & 0.112 {\tiny $\pm$ 0.03} & {0.428} {\tiny $\pm$ 0.01} & {0.403} {\tiny $\pm$ 0.02} & 0.542 {\tiny $\pm$ 0.00} & {0.338} {\tiny $\pm$ 0.01} & {0.318} {\tiny $\pm$ 0.02} & 0.248 {\tiny $\pm$ 0.04} & 0.290 {\tiny $\pm$ 0.05} \\
8ex & 0.169 {\tiny $\pm$ 0.03} & {0.430} {\tiny $\pm$ 0.03} & {0.331} {\tiny $\pm$ 0.03} & 0.570 {\tiny $\pm$ 0.01} & \textbf{0.429} {\tiny $\pm$ 0.05} & {0.337} {\tiny $\pm$ 0.01} & 0.200 {\tiny $\pm$ 0.04} & 0.339 {\tiny $\pm$ 0.18} \\
12ex & 0.148 {\tiny $\pm$ 0.03} & {0.453} {\tiny $\pm$ 0.02} & {0.384} {\tiny $\pm$ 0.02} & 0.565 {\tiny $\pm$ 0.01} & {0.410} {\tiny $\pm$ 0.06} & {0.412} {\tiny $\pm$ 0.03} & 0.160 {\tiny $\pm$ 0.02} & 0.351 {\tiny $\pm$ 0.08} \\
\hline
 \rowcolor{Gray}
\multicolumn{9}{l}{\textit{Experiments with Algorithmically Retrieved Examples}}
 \\
4ex $\textsc{BM25}_\textsc{r}$&  0.247 & 0.424 & 0.252 & 0.482 & 0.342 & 0.364 & 0.144 & 0.264 \\
4ex $\textsc{BM25}_\textsc{c}$&  0.129 & 0.424 & 0.339 & 0.510 & 0.370 & 0.172 & 0.192 & 0.549 \\
4ex $\textsc{BM25}_\textsc{c+r}$&  0.213 & 0.441 & 0.432 & 0.479 & 0.371 & 0.137 & 0.211 & 0.479 \\
8ex $\textsc{BM25}_\textsc{r}$&  0.309 & 0.487 & 0.275 & 0.536 & 0.304 & \textbf{0.426} & 0.121 & 0.419 \\
8ex $\textsc{BM25}_\textsc{c}$&  0.227 & 0.564 & 0.460 & 0.627 & 0.387 & 0.323 & 0.123 & 0.518 \\
8ex $\textsc{BM25}_\textsc{c+r}$& 0.185 & 0.458 & 0.439 & 0.526 & 0.308 & 0.377 & 0.171 & 0.530 \\
12ex $\textsc{BM25}_\textsc{r}$&  0.300 & 0.474 & 0.358 & 0.570 & 0.337 & 0.393 & {0.095*} & 0.414 \\
12ex $\textsc{BM25}_\textsc{c}$&  0.278 & \textbf{0.688}  & 0.449 & \textbf{0.674} & 0.397 & 0.377 & {0.106*} & 0.492 \\
12ex $\textsc{BM25}_\textsc{c+r}$& 0.202 & 0.491 & 0.452 & 0.465 & 0.349 & 0.358 & 0.148 & 0.493 \\
\hline
Best of DSTC10 baselines & \textbf{0.319} & 0.532 & \textbf{0.493} & 0.596 & 0.363 & 0.247 & \textbf{0.395} & \textbf{0.555} \\
% \hline
\end{tabular}
\caption{Spearman correlation of model predictions for overall quality with human ratings for TNLGv2 530B model with algorithmically chosen examples. TU, PU, PZ, DZ, CG, DGU, DGR, EG, FT and FD are abbreviations for
    TopicalChat-USR, PersonaChat-USR~~\cite{mehri-eskenazi-2020-usr},
    PersonaChat-Zhao~\cite{zhao-etal-2020-designing},
    DailyDialog-Zhao~\cite{zhao-etal-2020-designing}, ConvAI2-GRADE~\cite{huang-etal-2020-grade}, DailyDialog-Gupta~\cite{gupta-etal-2019-investigating}, DailyDialog-GRADE~\cite{huang-etal-2020-grade}, Empathetic-GRADE~\cite{huang-etal-2020-grade}, FED-Turn and FED-Dial~\cite{mehri-eskenazi-2020-unsupervised}.}
    \label{tab:eval530b}
    \vspace{-2mm}
  \label{table:dstc530b}
\end{table*}

\begin{table*}[t]
% \small
% \footnotesize
\scriptsize
\centering
\setlength\tabcolsep{7.0pt}
\begin{tabular}{lcccccccccccc}
    \hline
   Model & TU & DZ & PU & DGU & DGR & FT & EG & FD \\
   \hline
% BLOOM-7B-4ex & {0.022*}  & 0.07  & 0.113 & 0.122 & 0.125 & 0.097 & -0.052 & 0.277  \\
% OPT-6.7B-4ex  & {0.031*}  & 0.067  & \textbf{0.243} & 0.088 & 0.125 & 0.329 & -0.052 & 0.465  \\
% OPT-30B-4ex  & {-0.033*}  & \textbf{0.141*}  & -0.243 & \textbf{0.228} & \textbf{0.216} & \textbf{0.351} & \textbf{0.216*} & \textbf{0.764}  \\
% \hline
% FlantT5XL-3B-8ex & 0.378  & 0.629& 0.532  & 0.715  & 0.373  & 0.401  & 0.401  & 0.442 \\
% \hline
% FlantT5XL-3B-0ex & 0.367 & 0.594 & 0.525 & 0.674 & 0.359 & 0.361 & 0.410 & 0.389 \\
% InstructDial-3B-0ex  & 0.446  & 0.601  & 0.376 & 0.634 & 0.286 & 0.263 & 0.475 & 0.228  \\
\rowcolor{Gray}
\multicolumn{9}{l}{\textit{Few-shot in-context Experiments}}
 \\
BLOOM-7B-4ex & {0.027*}  & 0.075  & 0.123 & 0.127 & 0.131 & 0.117 & 0.012 & 0.289  \\
OPT-6.7B-4ex  & 0.115  & 0.258  & 0.444 & 0.228 & {0.091*} &0.486 & {0.044*} & \textbf{0.657}  \\
TNLG-6.7B-4ex &  0.124 & 0.198 & 0.237 & 0.209 & {0.214} & 0.296 & {0.057*} & 0.314\\
TNLG-530B-4ex &  0.129 & 0.424 & 0.339 & 0.510 & 0.370 & 0.172 & 0.192 & {0.549} \\
Flan-T5-3B-4ex & 0.447  & 0.657 & 0.578  & 0.714  & 0.379  & 0.442  & 0.396  & 0.492 \\
InstructGPT-175B-4ex  & \textbf{0.616} & \textbf{0.716} & \textbf{0.687} & \textbf{0.746} & \textbf{0.472} & 
\textbf{0.506} & 0.305 & 0.412  \\
\hline
\rowcolor{Gray}
\multicolumn{9}{l}{\textit{Zero-shot Experiments}}
 \\
Flan-T5-3B-0ex & 0.357 & 0.599 & 0.533 & 0.677 & 0.351 & 0.380 & 0.418 & 0.444 \\
InstructDial-3B-0ex  & 0.446  & 0.601  & 0.376 & 0.634 & 0.286 & 0.263 & \textbf{0.475} & 0.228  \\
% InstructDial-3B-8ex  & 0.350  & 0.362  & 0.379 & 0.416 & 0.221 & 0.302 & 0.086 & 0.239  \\
\hline
Best of DSTC10 baselines & 0.319 & 0.532 & 0.493 & 0.596 & 0.363 & 0.247 & 0.395 & 0.555 \\ 
%\qquad finetuned on dialogue & 0.319 & 0.532 & 0.493 & 0.596 & 0.363 & 0.247 & 0.395 & 0.555 \\
%(Finetuned on Dialogue) & & & & & & & & \\
%\qquad finetuned on DSTC 10 datasets & 0.319 & 0.532 & 0.493 & 0.596 & 0.254 & 0.247 & 0.395 & 0.555 \\
%(Finetuned on DSTC 10) & & & & & & & & \\
% \hline
Best TNLGv2 value & 0.309 & 0.688 & 0.460 & 0.678 & 0.429 & 0.426 & 0.248 & 0.549 \\
    \end{tabular}
    \caption{Spearman correlation of model predictions for overall quality with human ratings with 4 examples chosen with BM25 using context. Macro average scores are also shown.}
    \label{tab:evalother}
    \vspace{-2mm}
  \label{table:dstcother}
\end{table*}

Experiments are performed with randomly chosen examples and examples that were chosen by BM25 over 4, 8, and 12 examples in Table \ref{tab:eval530b} and Appendix \ref{sec:dstc106.7B}. Higher correlation results are obtained on 4 datasets (DZ, DGU, DGR, and FT) with comparable results on 3 datasets (TU, PU, and FD), as compared to the best DSTC10 baselines. Most of the best results are on the 530B TNLGv2 model, which will be discussed in this section, as compared to the 6.7B TNLGv2 model. Several factors are relevant here: the language of the dataset, the way the dataset was created, and how the dataset was annotated.

DailyDialog contains more formal language, thus TNLGv2 should perform well since its training dataset includes data sources with formal language. DZ, DGU, and DGR almost always perform the best when examples are chosen from looking at the context; adding the response generally leads to poorer performance. Since these datasets are annotated for \textit{appropriateness} and \textit{coherence}, context is more important than a more turn-specific metric.

TopicalChat was created through knowledge-grounding. The conversations could thus have more substance than a purely open-domain un-prompted conversation. It thus follows that response selection will work the best when choosing examples. PersonaChat has conversations that are persona-conditioned, so the quality of the conversation should take into account the entire conversation for each persona. It performs better with examples chosen for context and response or with just context.

FED is split into turn- and dialog-level annotations, thus, for turn-level annotations choosing examples based on responses should work best, and for dialog-level annotations choosing examples based on either the context or the context and response should perform the best. Choosing examples with context and response performs the best for EG, but randomly choosing examples outperforms that result. It may be that with emotionally grounded conversations, the model needs more, or more diverse examples due to the different ways emotion can be expressed. 

In general, choosing examples algorithmically improves performance over randomly choosing examples. This is consistent with previous experiments above. However, randomly-chosen examples perform better on the DGR and EG datasets on the 530B TNLGv2 model. This may be because these two datasets were rated for \textit{coherence}. Algorithmically, choosing examples based on context and response performs the best on EG, as was seen for coherence in FED in Section \ref{sec:incontext}. 

% DGR also either performs the best with choosing examples based on context only or with the context and response. 

\subsubsection{Comparisons Across LLMs}
Table~\ref{table:dstcother} compares the evaluation results across various LLMs.
Due to model input length restrictions, the following experiments were carried out using 4 in-context examples or in a zero-shot setting. BM25 is only used with the context as the example selection strategy, since it performed well with the TNLGv2 models.

In the few-shot setting, models that were not fine-tuned or trained with prompting (BLOOM, OPT) did not have consistent results across the datasets. However, those that were fine-tuned or prompted (Flan-T5, InstructGPT, InstructDial) had results that were close to or surpassed the previous best DSTC10 baselines. InstructGPT performed the best. Even in the zero-shot setting, Flan-T5 outperforms the baseline in 6 of the datasets, and InstructDial in 5.

These results clearly show that for dialog evaluation, it is insufficient to simply train on large amounts of general internet data. Specialized approaches such as instruction tuning on multiple tasks improve the generalization capabilities of models in zero- and few-shot settings. It is not surprising that InstructGPT performs the best since it fine-tunes a very large language model with instructions.

%% file: templates/sections/5_Conclusion.tex
\section{Conclusion}

% Large language models have the potential to contribute significantly to dialog evaluation. Out of the box LLMs can perform well for this task in a few-shot setting. However, this performance varies greatly depending on the wording and number of examples in the prompt. Performance is also affected by the data that a model is trained on. Smaller language models which are fine-tuned for prompting, trained on dialog data, and/or trained to perform dialog tasks outperform larger language models. This indicates that large language models should have more diverse pre-training data in order to be able to handle a larger variety of tasks. 

LLMs have the potential to significantly contribute to dialog evaluation. Current LLMs perform well for this task in a few-shot setting. However, this performance varies greatly depending on the content of and number of examples in the prompt. Models prefer more similar examples for metrics that they struggle to evaluate, while preferring examples with more diverse ratings for metrics that they can evaluate well. Very large language models also still afford performance gains, especially for overall quality evaluation at the turn and dialog level. Even though large language models perform better at dialog-level fine-grained metrics, there are still previously shown issues with how these models understand social situations and use context that may hinder further improvement if not addressed.

Performance is also affected by the model's training data. Smaller language models that are fine-tuned on instructions, trained on dialog data, and/or trained on multiple dialog tasks outperform larger language models. These smaller models also perform more consistently over different domains. This indicates that LLMs should have more diverse pre-training data in order to be able to handle a larger variety of tasks in few or zero-shot settings.

More work needs to be done on understanding how a large language model models different types of tasks. In-context example selection and example wording still remains unstable across large language models in many tasks, and the performance variation over different dialog domains in this paper demonstrates that as well. 

Presently, the LLMs explored in this paper have their own strengths. Smaller models such as BLOOM and OPT could share more training data similarity with dialog tasks based on their objective. TNLGv2 530B provides a very large language model that has shown improvement in dialog evaluation along with other NLP tasks. Flan-T5 and InstructDial show the efficacy of fine-tuning a LLM on dialog tasks, and InstructGPT shows the importance of training a model to better recognize prompts. The evaluations of these models provide suggestions for the characteristics of the best LLMs to use for dialog evaluation. Future work in using LLMs for other NLP tasks can benefit from such comprehensive analyses. Once a better understanding of LLMs is realized, the capabilities of large language models for zero- and few-shot tasks will increase greatly.

%% file: templates/sections/6_Acknowledgements.tex
\section{Acknowledgements}
We would like to thank Microsoft for allowing us to use TNLGv2. J.H. was supported by the NSF Graduate Research Fellowship under Grant Nos. DGE1745016 and DGE2140739. The opinions expressed in this paper do not necessarily reflect those of that funding agency.

%% file: templates/sections/7_Appendix.tex
\newpage
\appendix
\setcounter{section}{0}
\renewcommand\thesection{\Alph{section}}

\section{LLMs and Their Training/Fine-tuning Data}
\label{sec:llmdata}

\begin{table*}[h!]
\begin{center}
\small
\centering
\setlength\tabcolsep{1.3pt}
\begin{tabular}{ lccc } 
  \toprule
 & Seen Dialog & Fine-tuned \\
 \hline 
 Flan-T5 & \checkmark & \checkmark \\
 InstructDial & \checkmark & \checkmark \\
 InstructGPT & \checkmark & \checkmark \\
 BLOOM & \checkmark & $\times$\\
 OPT & \checkmark & $\times$\\
 TNLGv2 & $\times$ & $\times$\\
 %Flan-T5 & all the stuff from the paper & \checkmark & \checkmark & \checkmark \\
 %InstructDial & stuff from paper & \checkmark & \checkmark & \checkmark \\
 %InstructGPT & prompting dataset & idk & idk &  \checkmark \\
 %BLOOM & 46 languages and 13 programming languages & $\times$ & \checkmark & $\times$\\
 %OPT & RoBERTa, the Pile, PushShift.io Reddit & \checkmark & \checkmark & $\times$\\
 %TNLGv2 & the Pile, CommonCrawl snapshots, RealNews, CC-Stories & $\times$ & $\times$ & $\times$\\
 \bottomrule
\end{tabular}
\caption{LLMs with the datasets they were trained on. During training or fine-tuning: ``Seen Dialog" indicates that the model has explicitly seen dialog datasets and therefore elements of casual language, and ``fine-tuned" indicates that the model was fine-tuned on dialog data. TNLGv2 has not seen datasets explicitly categorized as having dialog, but elements of casual language may be included in the Common Crawl snapshots and other internet-based corpora. Symbols: \checkmark means that the category is included and $\times$ means that the category is not included.}
\label{table:appendixllm}
\end{center}
\end{table*}

\section{Prompt format examples FED}
\label{sec:exfed}

\begin{table*}[h!]
\begin{center}
\begin{tabular}{ |p{11cm}| } 
 \hline
 \textbf{Task}: Given a dialog history and a response, rate how interesting the response is with regards to the dialog history. \\
 \\ 
 \textbf{== Example 1 ==} \\
 A: Hi! \\
 B: Hi. This is a pleasant surprise. \\
 A: Haha...thanks! how did you like the gift? \\
 \textbf{Response}: Currently unpacking it I guess. How's your morning? \\
 \textbf{Rating}: 1/2 \\
 A: Hope you like it! Morning is good. Busy finishing up stuff before the holidays. \\
 B: I think I traveled too much the last couple of months so no holiday for me. But I'm okay with that. Going anywhere exciting? \\
 A: Yes \\
 \textbf{Response}: Where to? \\
 \textbf{Rating}: 1/2 \\
 A: Hawaii... looking forward to warm beaches. \\ 
 \textbf{Response}: WOW. Which island? I like Hawaii. \\
 \textbf{Rating}: 2/2 \\
 \hline
\end{tabular}
\caption{An example of a prompt with one example from FED \cite{mehri-eskenazi-2020-unsupervised}. Interestingness was rated in FED over three values corresponding to 0/2, 1/2, and 2/2. The resulting output is truncated to the integer value of 0, 1, or 2 to be used in evaluation.}
\end{center}
\end{table*}

\section{Prompt format examples DSTC10}
\label{sec:exdstc10}

\begin{table*}[h!]
\begin{center}
\begin{tabular}{ |p{11cm}| } 
 \hline
 \textbf{Instruction}: Given a conversation and a response, choose if the response is a good response to the context \\
 \\ 
 \textbf{Example} \\
 \textbf{Background info}: none \\
 \textbf{Conversation}:\\
 Person A: did your meal meet with your approval ?\\
 \textbf{Response}: yes , i did . it was a good meal . \\
 \textbf{Question}: Is the above response a good response to the conversation?\\
 \textbf{Answer}: Yes\\
 \\
 \textbf{Background info:} none \\
 \textbf{Conversation:} \\
 Person B: i really do hate public transportation.\\
 Person A: i agree , it 's just never on time.\\
 Response : you 're right.\\ 
 \textbf{Question}: Is the above response a good response to the conversation? \\
 \textbf{Answer}: \\
 \hline
\end{tabular}
\caption{An example of a prompt with examples from DSTC 10.}
\end{center}
\end{table*}

\section{Additional algorithmically chosen FED examples}
\label{sec:fedfull}

\begin{table*}[h!]
\begin{center}
\small
\centering
\setlength\tabcolsep{1.3pt}
\begin{tabular}{ lcccc } 
  \toprule
 & \multicolumn{2}{c}{$\textsc{BM25}_\textsc{c}$} & \multicolumn{2}{c}{$\textsc{BM25}_\textsc{r}$}  \\
 \cmidrule(lr){2-3}\cmidrule(lr){4-5}
 Quality & 7B & 530B & 7B & 530B \\
 \hline
 Interesting & 0.336 & 0.389 & 0.355 & 0.385  \\
 Engaging & 0.308 & 0.332 & 0.328 & 0.389 \\
 Specific & 0.217 & 0.224 & 0.297 & 0.329  \\
 Relevant & 0.338 & 0.314 & 0.311 & 0.356 \\
 Correct & 0.333 & 0.341 & 0.300 & 0.383 \\
 Sem. Approp. & 0.261 & 0.270 & 0.287 & 0.337 \\
 Understandable & 0.141 & 0.028* & 0.169 & 0.029* \\
 Fluent & 0.106 & 0.147 & 0.096* & 0.121 \\
 Overall & 0.435 & 0.438 & 0.360 & 0.407 \\
 \bottomrule
\end{tabular}
\caption{Turn-level fine-grained metrics on the FED dataset for algorithmically chosen examples over the TNLGv2 6.7B and 530B models. $\textsc{BM25}_\textsc{c}$ stands for examples chosen by BM25 considering the context and  $\textsc{BM25}_\textsc{r}$ stands for examples chosen by BM25 considering the response.}
\label{table:appendixc}
\end{center}
\end{table*}

\section{Additional LLM sizes on FED}
\label{sec:fedmodels}

\begin{table*}[h!]
\begin{center}
\small
\centering
\setlength\tabcolsep{1.3pt}
\begin{tabular}{ lcccc|cccc } 
  \toprule
 & \multicolumn{4}{c}{BLOOM} & \multicolumn{4}{c}{OPT} \\
 \cmidrule(lr){2-5}\cmidrule(lr){6-9}
 Quality & 560M & 1.1B & 1.7B & 3B & 125M & 350M & 1.3B & 2.7B \\
 \hline
 Interesting & 0.282 & 0.331 & 0.336 & 0.328 & 0.187 & 0.186 & 0.388 & 0.245 \\
 Engaging & 0.217 & 0.320 & 0.278 & 0.418 & 0.121 & 0.252 & 0.398 & 0.292 \\
 Specific & 0.030* & 0.065* & 0.204 & 0.353 & 0.197 & 0.004* & 0.217 & 0.222 \\
 Relevant & 0.076* & 0.056* & 0.072* & 0.091* & 0.146 & 0.105 & 0.231 & 0.177 \\
 Correct & 0.106 & 0.146 & 0.124 & 0.173 & 0.119 & 0.152 & 0.327 & 0.270 \\
 Sem. Approp. & 0.048* & 0.228 & 0.205 & 0.265 & 0.148 & 0.278 & 0.274 & 0.296 \\
 Understandable & -0.017* & 0.043* & -0.005* & 0.087* & 0.058* & 0.021* & 0.189 & 0.205 \\
 Fluent & 0.158 & \textbf{0.223} & 0.097* & 0.091* & 0.109 & 0.087* & 0.158 & 0.163 \\
 Overall & 0.086* & 0.179 & 0.076* & 0.285 & 0.134 & 0.219 & 0.338 & 0.197 \\
 \hline
\end{tabular}
\caption{Turn-level fine-grained metrics on the FED dataset for manually chosen examples over the smaller sizes of BLOOM and OPT.}
\label{table:appendixllmturn}
\end{center}
\end{table*}

\begin{table*}[h!]
\begin{center}
\small
\centering
\setlength\tabcolsep{1.3pt}
\begin{tabular}{ lcccc|cccc } 
 \hline
 & \multicolumn{4}{c}{BLOOM} & \multicolumn{4}{c}{OPT} \\
 \hline
 Quality & 560M & 1.1B & 1.7B & 3B & 125M & 350M & 1.3B & 2.7B \\
 \hline
 Coherent & 0.499 & 0.533 & 0.531 & 0.531 & 0.490 & 0.514 & 0.528 & 0.435 \\
 Error Recovery & 0.293 & 0.298 & 0.322 & 0.448 & 0.168 & 0.380 & 0.342 & 0.348 \\
 Consistent & 0.217 & 0.238 & 0.129* & 0.264 & 0.193 & 0.191 & 0.250 & 0.268 \\
 Diverse & 0.345 & 0.430 & 0.461 & 0.518 & 0.451 & 0.304 & 0.491 & 0.531 \\
 Topic Depth & 0.418 & 0.414 & 0.519 & 0.462 & 0.228 & 0.302 & 0.462 & 0.454 \\
 Likeable & 0.310 & 0.374 & 0.421 & 0.476 & 0.467 & 0.395 & 0.462 & 0.535 \\
 Understanding & 0.276 & 0.312 & 0.257 & 0.371 & 0.389 & 0.283 & 0.414 & 0.494 \\
 Flexible & 0.269 & 0.432 & 0.400 & 0.441 & 0.458 & 0.377 & 0.460 & 0.432 \\
 Informative & 0.149* & 0.384 & 0.372 & 0.537 & 0.378 & 0.402 & 0.381 & 0.544 \\
 Inquisitive & 0.198 & 0.350 & 0.318 & 0.339 & 0.489 & 0.300 & 0.439 & 0.413 \\
 Overall & 0.262 & 0.146* & 0.207 & 0.261 & -0.000* & 0.319 & 0.452 & 0.437 \\
 \hline
\end{tabular}
\caption{Dialog-level fine-grained metrics on the FED dataset for manually chosen examples over the smaller sizes of BLOOM and OPT.}
\label{table:appendixllmdialog}
\end{center}
\end{table*}

\newpage

\section{DSTC10 Results For TNLGv2 6.7B}
\label{sec:dstc106.7B}

\begin{table*}[h!]
% \small
% \footnotesize
\scriptsize
\centering
\setlength\tabcolsep{1.0pt}
\begin{tabular}{l|cccccccccccc}
    \hline
  Model & TU & DZ & PU & DGU & DGR & FT & EG & FD \\
  \hline
   \rowcolor{Gray}
\multicolumn{9}{l}{\textit{Experiments with Random Examples}}
 \\
4ex & {0.034*} {\tiny $\pm$ 0.05} & 0.117 {\tiny $\pm$ 0.02} & 0.206 {\tiny $\pm$ 0.02} & {0.080*} {\tiny $\pm$ 0.05} & {0.121}{\tiny $\pm$ 0.05} & 0.191 {\tiny $\pm$ 0.06} & {0.005*} {\tiny $\pm$ 0.04} & 0.228 {\tiny $\pm$ 0.03}\\
8ex & {0.054*} {\tiny $\pm$ 0.05} & 0.160 {\tiny $\pm$ 0.02} & 0.206 {\tiny $\pm$ 0.03} & {0.109*} {\tiny $\pm$ 0.03} & 0.139 {\tiny $\pm$ 0.08} & 0.178 {\tiny $\pm$ 0.02} & {0.060*} {\tiny $\pm$ 0.06} & 0.238 {\tiny $\pm$ 0.11} \\
12ex & {0.063*} {\tiny $\pm$ 0.03} & 0.149 {\tiny $\pm$ 0.00} & 0.225 {\tiny $\pm$ 0.01} & 0.114 {\tiny $\pm$ 0.05} & 0.143 {\tiny $\pm$ 0.06}  & 0.210 {\tiny $\pm$ 0.03} & {0.052*} {\tiny $\pm$ 0.02} & 0.127 {\tiny $\pm$ 0.04 } \\
\hline
   \rowcolor{Gray}
\multicolumn{9}{l}{\textit{Experiments with Algorithmically Retrieved Examples}}
 \\
4ex $\textsc{BM25}_\textsc{r}$ &  0.148 & 0.218 & 0.223 & 0.202 & {0.094*} & 0.273 & {-0.012*} & 0.335 \\
4ex $\textsc{BM25}_\textsc{c}$ &  0.124 & 0.198 & 0.237 & 0.209 & 0.214 & 0.296 & {0.057*} & 0.314\\
4ex $\textsc{BM25}_\textsc{c+r}$&  {0.05*} & 0.142 & 0.169 & 0.167 & {0.083*} & 0.274 & {0.038*} & 0.339 \\
8ex $\textsc{BM25}_\textsc{r}$ &  {0.077*} & 0.270 & 0.203 & 0.222 & 0.128 & 0.199 & {0.042*} & 0.335 \\
8ex $\textsc{BM25}_\textsc{c}$ &  0.184 & 0.328 & 0.343 & 0.526 & 0.176 & 0.363 & {0.073*} & 0.387\\
8ex $\textsc{BM25}_\textsc{c+r}$& {0.029*} & 0.152 & {0.020*} & 0.092 & {0.022*} & 0.348 & {0.024*} & 0.440 \\
12ex $\textsc{BM25}_\textsc{r}$ &  {0.069*} & 0.338 & 0.153 & 0.213 & {0.110*} & 0.250 & {0.026*} & 0.401 \\
12ex $\textsc{BM25}_\textsc{c}$ &  0.285 & 0.544 & 0.325 & 0.678 & 0.208 & 0.330 & {0.042*} & 0.365\\
12ex $\textsc{BM25}_\textsc{c+r}$& {0.035*} & 0.168 & {0.088*} & {0.086*} & {0.100*} & 0.407 & 0.092* & 0.343 \\
\end{tabular}
\caption{Spearman correlation of model predictions with human ratings for TNLGv2 6.7B model with algorithmically chosen examples. TU, PU, PZ, DZ, CG, DGU, DGR, EG, FT and FD are abbreviations for
    TopicalChat-USR, PersonaChat-USR~~\cite{mehri-eskenazi-2020-usr},
    PersonaChat-Zhao~\cite{zhao-etal-2020-designing},
    DailyDialog-Zhao~\cite{zhao-etal-2020-designing}, ConvAI2-GRADE~\cite{huang-etal-2020-grade}, DailyDialog-Gupta~\cite{gupta-etal-2019-investigating}, DailyDialog-GRADE~\cite{huang-etal-2020-grade}, Empathetic-GRADE~\cite{huang-etal-2020-grade}, FED-Turn and FED-Dial~\cite{mehri-eskenazi-2020-unsupervised}.}
    \label{tab:eval7b}
    \vspace{-2mm}
  \label{table:dstc7b}
\end{table*}

\section{DSTC10 Baseline Results}
\label{sec:dstc10}

\begin{table*}[h!]
\small
\centering
\setlength\tabcolsep{1.3pt}
% \begin{tabular}{lccccccccccccccc}
\begin{tabular}{lccrrrrrrrrrrrrrrr}
    \hline
    Model & Fine-Tuned on & TU & DZ & PU & DGU & DGR & FT & EG & FD \\
     & DTSC 10 datasets & & & & & & & & \\
    \hline
    % \checkmark & $\times$
% FlowScore~\cite{li-etal-2021-conversations} & \checkmark  & 0.068 & -0.063 & 0.053 & - & -0.043 & - & -0.009 \\
% DialogRPT~\cite{gao-etal-2020-dialogue} & \checkmark & 0.118 & 0.067 & -0.036 & 0.075 & 0.037 & -0.249 & 0.203 & -0.134 \\
% MAUDE~\cite{sinha-etal-2020-learning} & \checkmark & 0.136 & 0.120 & 0.306 & 0.192 & -0.073 & -0.11 & -0.057 & -0.285 \\
USL-H~\cite{phy-etal-2020-deconstruct} & \checkmark  & \textbf{0.319} & 0.385 & \textbf{0.493} & 0.481 & 0.09 & 0.115 & 0.237 & 0.202 \\
GRADE~\cite{huang-etal-2020-grade} & \checkmark & 0.176 & \textbf{0.532} & 0.329 & \textbf{0.596} & 0.254 & 0.048 & 0.300 & 0.106 \\
DynaEval~\cite{zhang-etal-2021-DynaEval} & \checkmark  & -0.013 & 0.169 & 0.148 & 0.038 & 0.122 & \textbf{0.247} & 0.159 & \textbf{0.555} \\
USR~\cite{mehri-eskenazi-2020-usr} & $\times$ & 0.291 & 0.363 & 0.140 & 0.353 & 0.066 & 0.055 & 0.268 & 0.084 \\
FED~\cite{mehri-eskenazi-2020-unsupervised} & $\times$ & -0.090 & -0.080 & -0.004 & 0.025 & -0.009 & 0.173 & 0.005 & 0.178 \\
% QuestEval~\cite{scialom-etal-2021-questeval} & $\times$ & 0.104 & 0.22 & 0.106 & 0.243 & -0.026 & 0.168 & 0.195 & 0.114 \\
DEB~\cite{sai2020improving} & $\times$ & 0.123 & 0.486 & 0.351 & 0.579 & \textbf{0.363} & 0.044 & \textbf{0.395} & 0.141 \\
\hline
Best & & 0.319 & 0.532 & 0.493 & 0.596 & 0.363 & 0.247 & 0.395 & 0.555 \\
\hline
    \end{tabular}
    \caption{Spearman correlation of model predictions with human ratings. The models fine-tuned on DSTC 10 datasets tend to perform better on the DSTC 10 datasets. TU, PU, PZ, DZ, CG, DGU, DGR, EG, FT and FD are abbreviations for
    TopicalChat-USR, PersonaChat-USR~~\cite{mehri-eskenazi-2020-usr},
    PersonaChat-Zhao~\cite{zhao-etal-2020-designing},
    DailyDialog-Zhao~\cite{zhao-etal-2020-designing}, ConvAI2-GRADE~\cite{huang-etal-2020-grade}, DailyDialog-Gupta~\cite{gupta-etal-2019-investigating}, DailyDialog-GRADE~\cite{huang-etal-2020-grade}, Empathetic-GRADE~\cite{huang-etal-2020-grade}, FED-Turn and FED-Dial~\cite{mehri-eskenazi-2020-unsupervised}.}
    \label{tab:eval}
    \vspace{-2mm}
\end{table*}

%% file: editor/templates/appendix.tex
%%%%%%%%%%%%%%%%%%%%% appendix.tex %%%%%%%%%%%%%%%%%%%%%%%%%%%%%%%%%
%
% sample appendix
%
% Use this file as a template for your own input.
%
%%%%%%%%%%%%%%%%%%%%%%%% Springer-Verlag %%%%%%%%%%%%%%%%%%%%%%%%%%

\chapter{Chapter Heading}
\label{introA} % Always give a unique label
% use \chaptermark{}
% to alter or adjust the chapter heading in the running head

Use the template \emph{appendix.tex} together with the Springer document class SVMono (monograph-type books) or SVMult (edited books) to style appendix of your book in the Springer layout.

\section{Section Heading}
\label{sec:A1}
% Always give a unique label
% and use \ref{<label>} for cross-references
% and \cite{<label>} for bibliographic references
% use \sectionmark{}
% to alter or adjust the section heading in the running head
Instead of simply listing headings of different levels we recommend to let every heading be followed by at least a short passage of text. Further on please use the \LaTeX\ automatism for all your cross-references and citations.

\subsection{Subsection Heading}
\label{sec:A2}
Instead of simply listing headings of different levels we recommend to let every heading be followed by at least a short passage of text. Further on please use the \LaTeX\ automatism for all your cross-references and citations as has already been described in Sect.~\ref{sec:A1}.

For multiline equations we recommend to use the \verb|eqnarray| environment.
\begin{eqnarray}
\vec{a}\times\vec{b}=\vec{c} \nonumber\\
\vec{a}\times\vec{b}=\vec{c}
\label{eq:A01}
\end{eqnarray}

\subsubsection{Subsubsection Heading}
Instead of simply listing headings of different levels we recommend to let every heading be followed by at least a short passage of text. Further on please use the \LaTeX\ automatism for all your cross-references and citations as has already been described in Sect.~\ref{sec:A2}.

Please note that the first line of text that follows a heading is not indented, whereas the first lines of all subsequent paragraphs are.

% For figures use
%
\begin{figure}[t]
\sidecaption[t]
% Use the relevant command for your figure-insertion program
% to insert the figure file.
% For example, with the graphicx style use
\includegraphics[scale=.65]{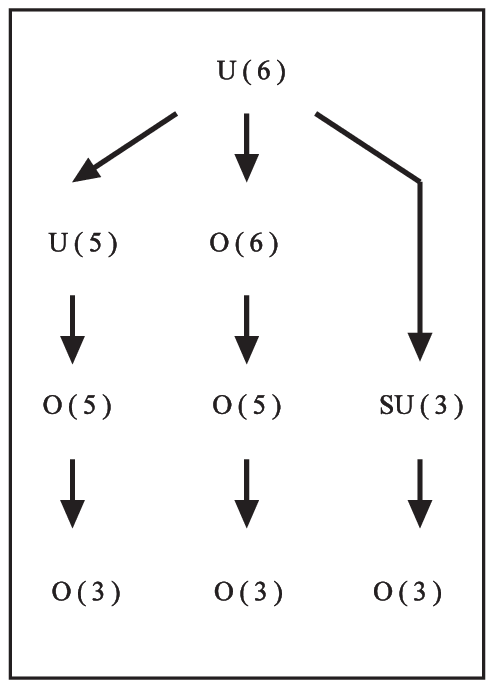}
%
% If no graphics program available, insert a blank space i.e. use
%\picplace{5cm}{2cm} % Give the correct figure height and width in cm
%
\caption{Please write your figure caption here}
\label{fig:A1}       % Give a unique label
\end{figure}

% For tables use
%
\begin{table}
\caption{Please write your table caption here}
\label{tab:A1}       % Give a unique label
%
% Follow this input for your own table layout
%
\begin{tabular}{p{2cm}p{2.4cm}p{2cm}p{4.9cm}}
\hline\noalign{\smallskip}
Classes & Subclass & Length & Action Mechanism  \\
\noalign{\smallskip}\hline\noalign{\smallskip}
Translation & mRNA$^a$  & 22 (19--25) & Translation repression, mRNA cleavage\\
Translation & mRNA cleavage & 21 & mRNA cleavage\\
Translation & mRNA  & 21--22 & mRNA cleavage\\
Translation & mRNA  & 24--26 & Histone and DNA Modification\\
\noalign{\smallskip}\hline\noalign{\smallskip}
\end{tabular}
$^a$ Table foot note (with superscript)
\end{table}
%

%% file: editor/templates/glossary.tex
%%%%%%%%%%%%%%%%%%%%%%acronym.tex%%%%%%%%%%%%%%%%%%%%%%%%%%%%%%%%%%%%%%%%%
% sample list of acronyms
%
% Use this file as a template for your own input.
%
%%%%%%%%%%%%%%%%%%%%%%%% Springer %%%%%%%%%%%%%%%%%%%%%%%%%%

\Extrachap{Glossary}

Use the template \emph{glossary.tex} together with the Springer document class SVMono (monograph-type books) or SVMult (edited books) to style your glossary\index{glossary} in the Springer layout.

\runinhead{glossary term} Write here the description of the glossary term. Write here the description of the glossary term. Write here the description of the glossary term.

\runinhead{glossary term} Write here the description of the glossary term. Write here the description of the glossary term. Write here the description of the glossary term.

\runinhead{glossary term} Write here the description of the glossary term. Write here the description of the glossary term. Write here the description of the glossary term.

\runinhead{glossary term} Write here the description of the glossary term. Write here the description of the glossary term. Write here the description of the glossary term.

\runinhead{glossary term} Write here the description of the glossary term. Write here the description of the glossary term. Write here the description of the glossary term.